\documentclass[letterpaper, 10 pt, conference]{ieeeconf}
\IEEEoverridecommandlockouts
\let\labelindent\relax
\usepackage{enumitem}
\usepackage{balance}
\usepackage[labelformat=simple]{subfig}
\usepackage[font={footnotesize}]{caption}
\usepackage{array}
\usepackage{textcomp}
\usepackage{mathtools, nccmath}
\usepackage{graphicx}
\usepackage{amsfonts}
\usepackage{amsmath}
\usepackage{autobreak}
\usepackage{amssymb}
\usepackage{tikz}
\usepackage{arydshln}
\usepackage{multirow}
\usepackage{bm}
\usepackage{epstopdf}
\usepackage{cite}
\usepackage{siunitx}
\usepackage{xcolor}
\usepackage{float} 
\usepackage{stfloats}
\usepackage{lineno}
\usepackage{subfloat}
\usepackage{colortbl}
\usepackage{booktabs}
\usepackage[ruled,linesnumbered,boxed,noend]{algorithm2e}

\makeatletter
\let\NAT@parse\undefined
\makeatother
\usepackage[colorlinks=true,
            linkcolor=blue,
           ]{hyperref}

\allowdisplaybreaks[4]

\title{\LARGE \bf Data-Driven MPC with Data Selection for\\
Flexible Cable-Driven Robotic Arms}
\author{Huayue Liang$^*$, Yanbo Chen$^*$, Hongyang Cheng, Yanzhao Yu, Shoujie Li, \\Junbo Tan, Xueqian Wang$^\dagger$, and Long Zeng$^\dagger$
\thanks{$*$ indicates equal contribution.}
\thanks{$^\dagger$ Corresponding authors: Xueqian Wang, Long Zeng.}
\thanks{This work was supported by the Natural Science Foundation of Shenzhen (No.JCYJ20230807111604008, No. JCYJ20240813112007010), the Natural Science Foundation of Guangdong Province (No.2024A1515010003), National Key Research and Development Program of China (No. 2022YFB4701400) and the Shenzhen Major Undertaking Plan (CJGJZD20240729141702003).}
\thanks{Huayue Liang, Yanbo Chen, Hongyang Cheng, Yanzhao Yu, Shoujie Li, Junbo Tan and Xueqian Wang are with the Center for Artificial Intelligence and Robotics, Shenzhen International Graduate School, Tsinghua University, Shenzhen 518055, China,{\tt\footnotesize\{lianghy23@mails,
cyb23@mails,
chenghy22@mails,
yuyz24@mails,
lsj20@mails,
tjblql@sz,
wang.xq@sz\}.tsinghua.edu.cn}.}
\thanks{Long Zeng is with the Department of Advanced Manufacturing, Shenzhen International Graduate School, Tsinghua University, Shenzhen 518055, China, 
\tt {zenglong@sz.tsinghua.edu.cn}.}
}

\begin{document}

\maketitle
\begin{abstract}


Flexible cable-driven robotic arms (FCRAs) offer dexterous and compliant motion. Still, the inherent properties of cables, such as resilience, hysteresis, and friction, often lead to particular difficulties in modeling and control. This paper proposes a model predictive control (MPC) method that relies exclusively on input-output data, without a physical model, to improve the control accuracy of FCRAs. First, we develop an implicit model based on input-output data and integrate it into an MPC optimization framework. Second, a data selection algorithm (DSA) is introduced to filter the data that best characterize the system, thereby reducing the solution time per step to approximately 4 ms, which is an improvement of nearly 80\%. Lastly, the influence of hyperparameters on tracking error is investigated through simulation. The proposed method has been validated on a real FCRA platform, including five-point positioning accuracy tests, a five-point response tracking test, and trajectory tracking for letter drawing. The results demonstrate that the average positioning accuracy is approximately 2.070 mm. Moreover, compared to the PID method with an average tracking error of 1.418°, the proposed method achieves an average tracking error of 0.541°.

\end{abstract}

\section{Introduction}
\label{sec:Introduction}

Flexible cable-driven robotic arms (FCRAs) represent an innovative design\cite{gravagne2002manipulability,xu2008investigation}. These systems utilize the flexibility and extensibility of cables to achieve joint actuation, significantly enhancing the arm's flexibility and operational range. However, the materials used for cables\cite{porto2019position,baek2020hysteresis,luo2025d3} have inherent limitations, such as resilience, hysteresis, and friction, making the dynamic behavior of the arms complex and difficult to predict during motion. This poses significant challenges for control system design. Traditional control methods often rely on physical modeling but usually fail to capture nonlinear characteristics accurately, resulting in low control precision, slow response, and poor stability for high-precision tasks\cite{li2022design}.

To address these challenges, we propose a data-driven implicit modeling approach\cite{berberich2020data,coulson2021distributionally,bongard2022robust}. This method is based on behavioral systems theory and utilizes real-world data to uncover the motion patterns of FCRA. Unlike traditional methods, it does not rely on physical parameters. Instead, it derives dynamic characteristics directly from the input-output data, effectively overcoming the modeling difficulties caused by the properties of the cable material\cite{wang2018practical}.

We integrate this implicit model into a model predictive control (MPC) framework. The framework employs optimization algorithms to generate optimal motion trajectories. Additionally, we introduce a data selection algorithm (DSA) that extracts data more accurately reflecting the system's characteristics, further enhancing the effectiveness of MPC optimization. That is, the optimization time for each step is reduced from 19 to 4 ms.

The validation results confirm the advantages of the proposed strategy. The data-driven MPC approach significantly improves control and planning performance, achieving better trajectory tracking accuracy and greater system stability than traditional methods.

\begin{figure}
    \centering
    \includegraphics[width=8.5cm]{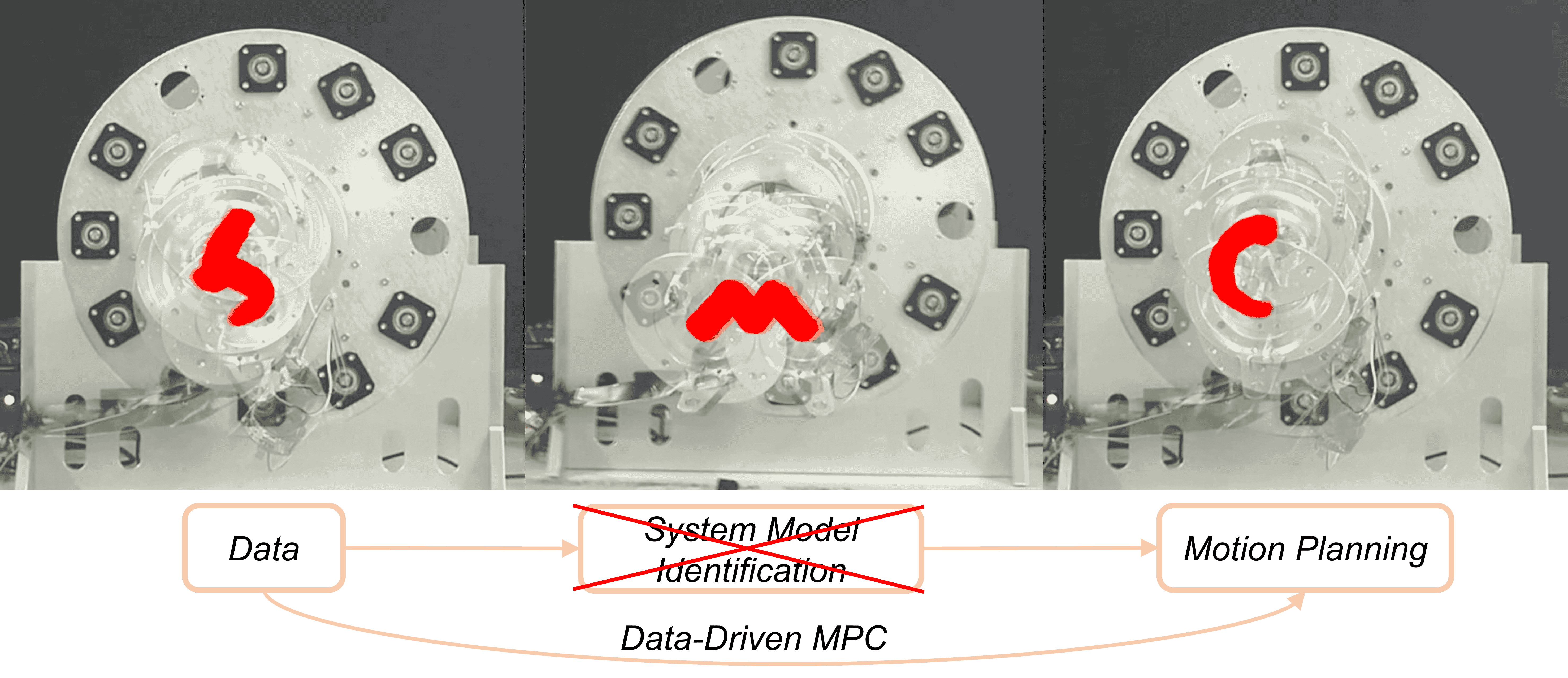}
    \caption{
The FCRA successfully follows a predefined trajectory using our data-driven MPC approach without a physical model, effectively illustrating the letters ``SMC".
    }
    \label{fig:SMC}
    \vspace{-0.2cm}
\end{figure}


This work offers the following contributions:
\begin{itemize}
\item This study overcomes the modeling challenges of the FCRA by proposing a data-driven MPC approach based exclusively on input-output data. The method utilizes extensive trajectory samples to uncover motion patterns, enabling precise behavioral predictions.
\item This study proposes a DSA that extracts the input-output data most representative of the system's characteristics from multiple system samples. It reduces the solving time of each optimization to approximately 4 ms, representing an improvement of nearly 80\%.
\item This study investigates the influence of hyperparameters on tracking error and validates the proposed algorithm on a real FCRA experimental platform. The results demonstrate that the average positioning accuracy of the five points is approximately 2.070 mm. Furthermore, compared to the PID method with an average tracking error of 1.418°, the proposed method achieves an average tracking error of 0.541°.
\end{itemize}

\section{Related Work}
\label{sub:related}
In recent years, FCRAs have emerged as a prominent research topic due to their lightweight and flexible characteristics. These have garnered significant attention across various applications, such as medical robotics, space exploration, and industrial automation. These attributes enable FCRAs to navigate complex environments and perform tasks that are difficult for rigid-bodied robots\cite{gravagne2002manipulability,xu2008investigation,porto2019position,baek2020hysteresis,luo2025d3}. However, controlling and planning such systems remain challenging. Although existing studies have made progress in structural design and control strategies\cite{gravagne2002manipulability,hannan2003kinematics,camarillo2009task}, most approaches are based on the construction of complex physical models. Due to the nonlinear properties of cable materials, such as friction and hysteresis effects\cite{porto2019position}, these physical models often struggle to capture the system's dynamic behavior accurately. Consequently, control performance is usually suboptimal and is below the requirements for high-precision tasks. There is a pressing need for an effective control method to enhance their performance and flexibility.

MPC is widely recognized as a robust framework for planning and control\cite{qin1997overview,jian2023dynamic,chen2023quadruped}. But building an accurate physical model of cables and their dynamic systems is difficult, and ensuring precision is challenging. Therefore, bypassing complex nonlinear modeling\cite{wang2018practical} and directly using input-output data from experiments for predictive control is an excellent approach. This method avoids the inaccuracies associated with traditional modeling while leveraging real-world data to capture the system's behavior more effectively. An MPC method, based exclusively on input-output data, has been proven effective in systems such as quadruped robots, aerial robots, and oil tanks\cite{fawcett2022toward,coulson2019data,berberich2021data,schmitt2023data,huang2019data}. Applying this approach to FCRAs is a logical next step. This method allows for precise control and planning without a nonlinear model, enabling tasks like trajectory following and dynamic environment interaction. Model-free control methods, such as reinforcement learning\cite{xiong2020comparison,yang2023manipulability,10160491}, are becoming increasingly popular. However, their practicality in physical systems remains limited due to poor performance and extensive training requirements.
This paper focuses on adapting and applying the MPC method to FCRAs, to achieve optimal control and trajectory following. 

We will explore how to tailor the MPC framework to the specific characteristics of FCRAs, including their unique dynamics and operational constraints, and demonstrate its effectiveness through simulations and experiments. This approach is our study's central focus as it balances computational efficiency, accuracy, and adaptability.

\section{Overview}
\label{sec:Overview}

\begin{figure*}[!ht]
    \centering
    \includegraphics[width=17cm]{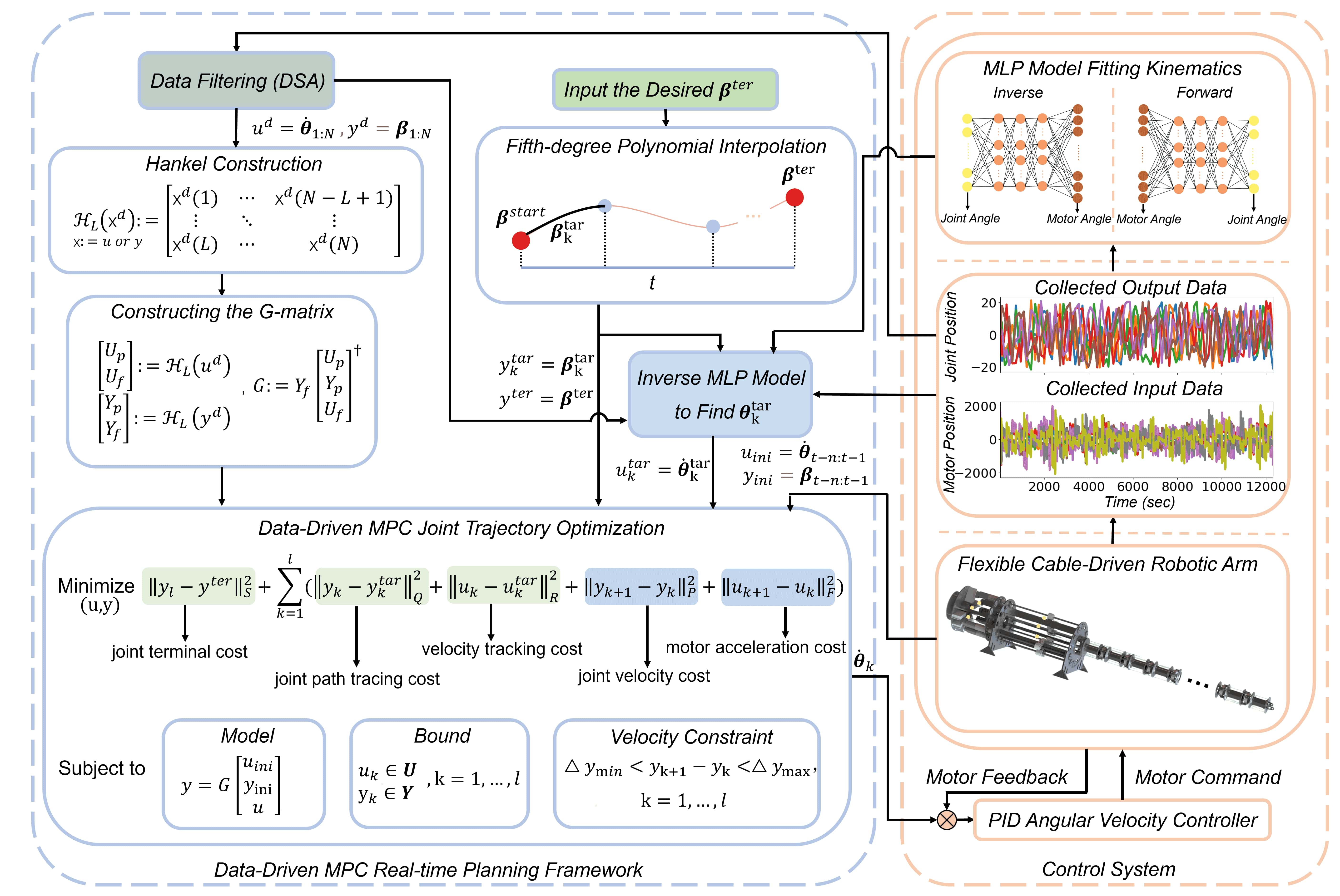}
    \caption{
        Overview of the System Framework.
        The system consists of two main components: the control system and the data-driven MPC real-time planning framework. The control system models an $n$-segment robotic arm, with forward and inverse MLP models constructed using \(N_s\) input-output pairs. The framework employs a data-driven MPC approach to achieve real-time planning.
    }
    \label{fig_framework}
\end{figure*}

The system framework illustrated in Fig.~\ref{fig_framework} is designed to enable real-time planning and control for FCRAs. It consists of the control system and the data-driven MPC real-time planning framework. The control system models an $n$-segment robotic arm, which integrates a PID motor angular velocity controller. The experimental model is initially introduced in Sec.~\ref{subsec:Modeling}. \(N_s\) input-output pairs, consisting of motor and joint angles, are collected. Using the multi-layer perceptron (MLP), the input-output data are fitted to construct both forward and inverse models, as described in Sec.~\ref{subsec:Equivalent Model Reference}. The forward model is used for algorithm validation, while the inverse model is employed to compute the reference input at each step.

Subsequently, the data-driven MPC real-time planning framework employs a method based on the collected data, as outlined in Sec.~\ref{subsec:Construction of Data-Driven MPC}. These data are filtered by an algorithm, as described in Sec.~\ref{subsec:Data Search Algorithm}. These data are utilized to construct a Hankel matrix and a $G$ matrix. The $G$ matrix is incorporated into initial constraints and the MPC planning is performed using a cost function, as detailed in Sec.~\ref{subsec:costfunction}. The angles of the terminal and target joint are derived from quintic polynomial interpolation, as explained in Sec.~\ref{subsec:Quintic polynomial interpolation}. The angles of the $L$-step target joint are fed into the inverse model in Sec.~\ref{subsec:Equivalent Model Reference}, which computes the corresponding target inputs. During each planning step, the system performs real-time calculations and feedback to update motor velocities and joint angles.

\section{Methods} \label{sec:Methods}

\subsection{Modeling}
\label{subsec:Modeling}
\begin{figure}[]
    \centering
    \includegraphics[width=8.65cm]{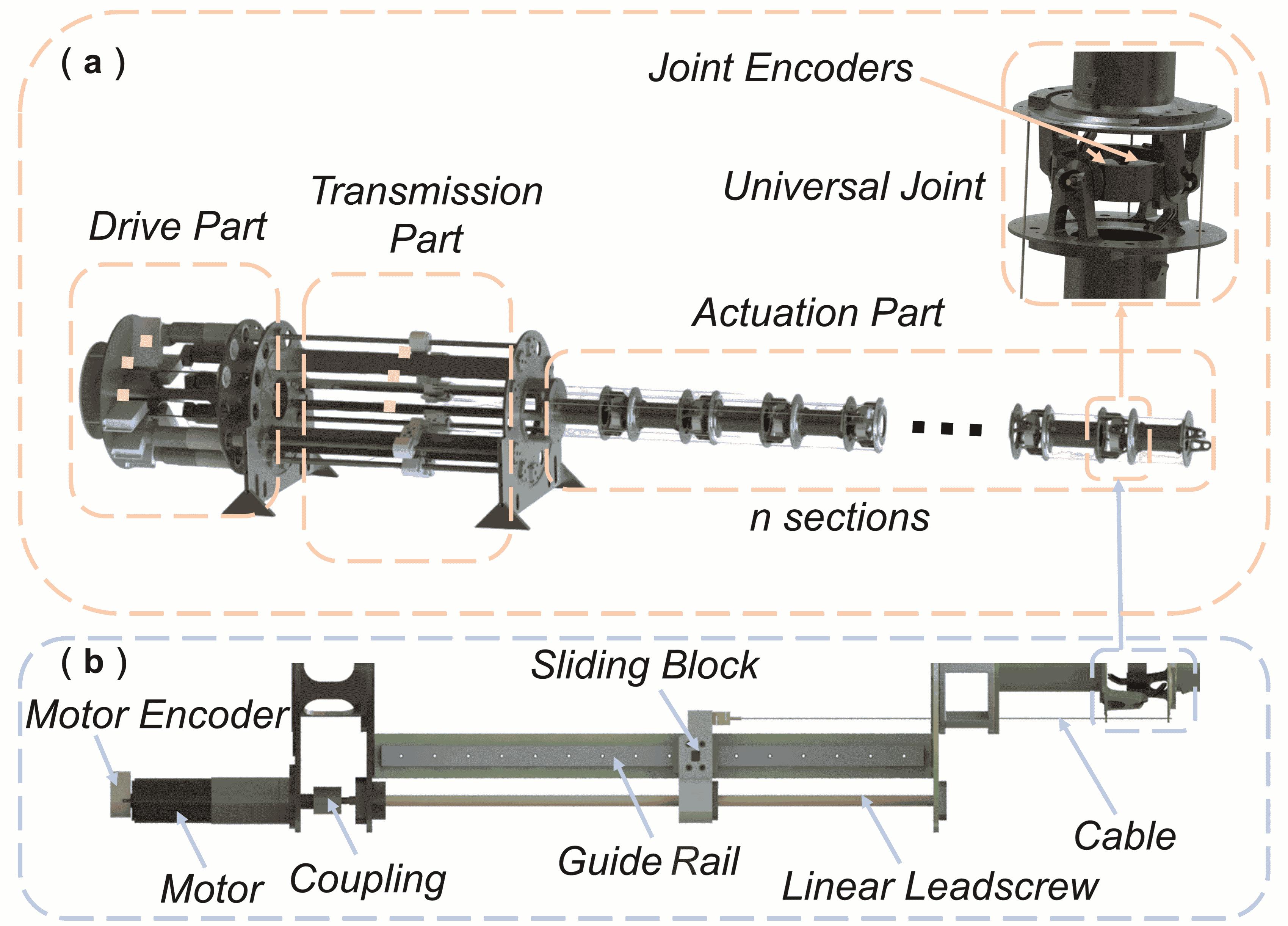}
    \caption{
        Structural composition of the FCRA. (a) illustrates that the FCRA is classified into three functional parts: the drive, the transmission, and the actuation parts. (b) describes the mechanical transmission characteristics of the mechanism. 
    }
    \label{fig:biaoshi}
\end{figure}
The multi-segment FCRA, shown in Fig.~\ref{fig:biaoshi}(a), consists of three primary parts and comprises $n$ segments:

\emph{1) Drive part}: This part comprises $3n$ brushless DC servo motors, each equipped with an encoder to measure its angle in real time. Each motor has its driver and receives commands via the CANOPEN protocol from the PC.

\emph{2) Transmission part}: This part includes linear leadscrews, couplings, and guide rails associated with the motors. The linear motion of the sliding blocks, which are threaded for precise movement, pulls the cables, thereby driving the joint movements, as illustrated in Fig.~\ref{fig:biaoshi}(b).

\emph{3) Actuation part}: This part consists of multiple universal joints. Each joint is driven by three cables and has two perpendicular degrees of freedom. Two encoders are installed on each joint to measure its angles.

\begin{figure}
    \centering
    \includegraphics[width=8.65cm]{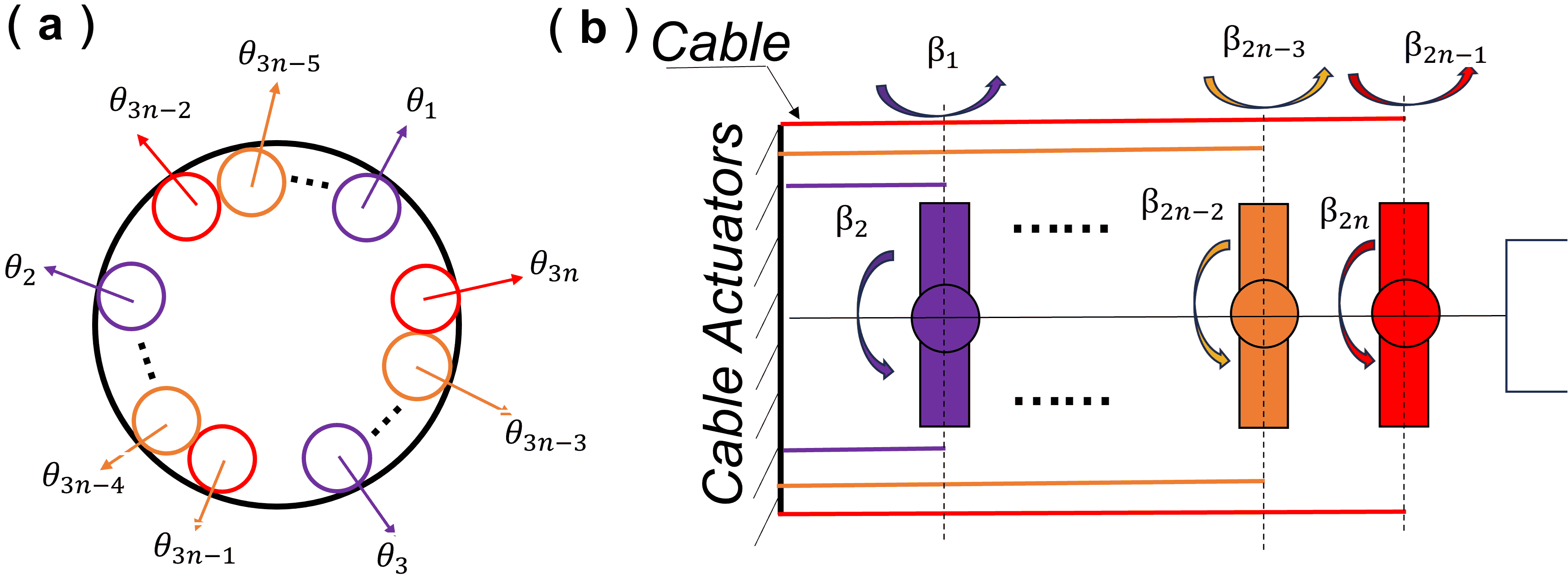}
    \caption{
         Motor arrangement. (a) illustrates the arrangement of each motor and its angle labels. Three motors, sharing the same color and connected by three cables, collectively drive a universal joint. (b) illustrates that the angles of the $3n$ motors are mapped to $2n$ sets of joint angles.
    }
    \label{fig:fuhao}
\end{figure}

The arrangement of the $3n$ motors is shown in Fig.~\ref{fig:fuhao}(a). Motors of the same color form a group, with their angles labeled counterclockwise as $\mathbf{\theta}_{1}$ through $\mathbf{\theta}_{{3n}}$ in three complete circles. Each motor group drives the rotation of a linear leadscrew. The rotation of the leadscrews stretches the cables, moving the universal joints and changing the two mutually perpendicular rotation angles, as shown in Fig.~\ref{fig:fuhao}(b). The rotation angles of the three universal joints are denoted as $\mathbf{\beta}_{1}$ through $\mathbf{\beta}_{{2n}}$.

However, a coupling phenomenon occurs in the FCRA. Any motion in the two degrees of freedom of the joints near the motors affects the length of the driving cables at the end-effector joints. Therefore, it cannot be assumed that the movements of the three motors directly correspond to changes in the two joints. Removal of joints requires complex formulas. Instead, we model the system as a black-box model, where the input angles of the $3n$ motors are represented as: 
\[
\bm{\theta} = \left( \mathbf{\theta}_{{1}}, \mathbf{\theta}_{{2}}, \mathbf{\theta}_{{3}}, \dots, \mathbf{\theta}_{{3n}} \right)^\top \in \mathbb{R}^{3n},
\] 
which correspond to the output angles of the $2n$ joints: 
\[
\bm{\beta} = \left( \mathbf{\beta}_{{1}}, \mathbf{\beta}_{{2}}, \dots, \mathbf{\beta}_{{2n}} \right)^\top \in \mathbb{R}^{2n}.
\]

\subsection{Construction of Data-Driven MPC}
\label{subsec:Construction of Data-Driven MPC}

For a linear time-invariant (LTI) system model, its state-space equations can be represented as:
\begin{equation}
    \begin{aligned}
    x_{t+1} &= Ax_{t} + Bu_{t} \\
    y_{t} &= Cx_{t} + Du_{t}
    \label{equ:system}
    \end{aligned}
\end{equation}
where $A \in \mathbb{R}^{h\times h}$, \, $B \in \mathbb{R}^{h\times m}$, \, $C \in \mathbb{R}^{c\times h}$, \, \text{and} \, $D \in \mathbb{R}^{c\times m}$ are spatial matrices of certain dimensions. \(h\), \(m\), and \(c\) represent the number of states, inputs, and outputs, respectively. 

We now define some terms: \(N \) denotes the total size of the collected data, \(l\) represents the prediction horizon, \(n_{\text{ini}}\) indicates the estimation horizon and a signal sequence composed of the collected data \(u^{d}\) and \(y^{d}\), where $u^{d} = \{u^{d}_{1:N}\} $ and $y^{d} = \{y^{d}_{1:N}\} $. And we have 
\begin{equation}
u^d_{a:b} = \begin{bmatrix}
u^d_{a} & \cdots & u^d_{b}
\end{bmatrix}^T
\label{duide}
\end{equation}
Similarly, we define \(y^d_{a:b}\). 
We denote the input and output sequences from \(a\) to \(b\) by \(u^{d}_{a:b}\) and \(y^{d}_{a:b}\), respectively. For this input trajectory signal, a Hankel matrix with \(L \) rows can be constructed as follows:
\begin{equation}
\mathcal{H}_L(u^{d}) := 
\begin{bmatrix}
u^{d}_1 & u^{d}_2 & \cdots & u^{d}_{N-L+1} \\
u^{d}_2 & u^{d}_3 & \cdots & u^{d}_{N-L+2} \\
\vdots & \vdots & \ddots & \vdots \\
u^{d}_L & u^{d}_{L+1} & \cdots & u^{d}_{N}
\end{bmatrix}
\label{equ:hankel}
\end{equation}

\textbf{Definition 1(\!\cite{berberich2020data}):} If \(\mathcal{H}_L(u^d)\) is the rank of full row , the signal \(u^d\) can be called a persistently exciting sequence of order \(L\). Meanwhile, the signal sequence satisfying Definition 1 must be sufficiently long to ensure that the system's state can be fully observed. The size of \(N\) must satisfy \(N \geq (m+1)L - 1\).

\textbf{Definition 2(\!\cite{berberich2020data}):} If there exists an initial state \(x_{0}\) that satisfies the state and output equations in (\ref{equ:system}), then the sequence $\{u^d_{1:N}, y^d_{1:N}\}$ is a trajectory of (\ref{equ:system}).

\textbf{Theorem 1(\!\cite{willems2005note}):} If $\{u^d_{1:N}, y^d_{1:N}\}$ is a trajectory of (\ref{equ:system}), and $u^d$ is persistently exciting of order $L+h$, then $\{u^d_{1:L}, y^d_{1:L}\}$ is a trajectory of the system if and only if there exists a vector $K \in \mathbb{R}^{N-L+1}$ such that
\begin{equation}
    \begin{bmatrix}
    \mathcal{H}_L(u^d) \\
    \mathcal{H}_L(y^d)
    \end{bmatrix}
    K =
    \begin{bmatrix}
    u \\
    y
    \end{bmatrix}
    \label{equ:theorem1}
\end{equation}

The Hankel matrix described in Theorem 1 can represent the trajectories of an unknown LTI system without explicit system identification or any deep learning processes. This theorem will be applied in Sec.~\ref{subsec:costfunction} for real-time motion planning of an FCRA.

We consider two distinct horizons contained within \(L = l + n_{\text{ini}}\): the estimation horizon \(n_{\text{ini}}\), which represents the number of input-output pairs required to uniquely determine the initial conditions of the given sequence $\left\{ u_{1:L}, y_{1:L} \right\}$ in (\ref{equ:theorem1}), and the prediction horizon \(l\), as commonly used in traditional MPC. 
After selecting the data $\{u^d_{1:N}, y^d_{1:N}\}$ using the DSA described in Sec.~\ref{subsec:Data Search Algorithm}, the hankel matrix in (\ref{equ:theorem1}) can be decomposed as follows:
\begin{equation}
    \begin{bmatrix} U_p \\ U_f \end{bmatrix} := \mathcal{H}_L(u^d), \quad  \begin{bmatrix} Y_p \\ Y_f \end{bmatrix}:= \mathcal{H}_L(y^d)
    \label{equ:hankel_decom}
\end{equation}
where \( U_p \in \mathbb{R}^{mn_{\text{ini}} \times (N-L+1)} \) and \( Y_p \in \mathbb{R}^{cn_{\text{ini}} \times (N-L+1)} \) represent the past input and output data used to estimate the initial condition, respectively. Meanwhile, \( U_f \in \mathbb{R}^{ml \times (N-L+1)} \) and \( Y_f \in \mathbb{R}^{cl \times (N-L+1)} \) represent future data used for prediction. The condition for the dataset size \( N \) is that it must satisfy \( N \geq (m+1)(L+h) - 1 \).

As noted in Sec.~\ref{sec:Introduction}, the dynamics of the FCRA does not satisfy the LTI assumption. The input-output data used in (\ref{equ:hankel}) originate from a nonlinear system and are subject to noise, which causes the Hankel matrix to misalign with possible system trajectories. 
Additionally, the online measurement values \( y_{t-k} \), used to initialize the prediction trajectory, are also influenced by measurement noise, leading to inaccurate predictions. 
Therefore, these inferences for linear systems do not necessarily hold for nonlinear systems.

When the data volume \( N \) is large, the method can be effectively extended to nonlinear systems.
Given that the size of \( K \) is positively correlated with \( N \), it is necessary to consider a least squares approximation of \( K \). The least squares method estimates \( K \), so it can be effectively eliminated from the problem, thus facilitating a constant linear mapping based exclusively on input-output data. Referring to \cite{huang2019data}, the least-squares approximation of \( K \) can be transformed into the following offline optimization problem:
\begin{equation}
    \begin{aligned}
    & \min_{K} \| K \|^2 \\
    & \text{s.t.} \quad 
    \begin{bmatrix}
    U_p \\
    Y_p \\
    U_f
    \end{bmatrix}
    K = 
    \begin{bmatrix}
    u_{\text{ini}} \\
    y_{\text{ini}} \\
    u
    \end{bmatrix}.
    \end{aligned}
    \label{equ:euq_k_MIN}
    \end{equation}
    
    The variables $u_{\text{ini}}$ and $y_{\text{ini}}$ represent the past trajectories over the time horizon $n_{\text{ini}}$. The variables $u$ and $y$ denote the predicted trajectories over the time horizon $l$. (\ref{equ:euq_k_MIN}) can be reformulated as:
\begin{equation}
    K = 
    \begin{bmatrix}
    U_p \\
    Y_p \\
    U_f
    \end{bmatrix}^\dagger
    \begin{bmatrix}
    u_{\text{ini}} \\
    y_{\text{ini}} \\
    u
    \end{bmatrix}
\end{equation}
where \((\cdot)^\dagger\) represents the pseudo-inverse. Based on \(y = Y_f K\), we have:
\begin{equation}
    y = G 
    \begin{bmatrix}
    u_{\text{ini}} \\
    y_{\text{ini}} \\
    u
    \end{bmatrix}, \quad
    G := Y_f 
    \begin{bmatrix}
    U_p \\
    Y_p \\
    U_f
    \end{bmatrix}^\dagger
    \label{equ:euq_g}
\end{equation}
where \(G\) can be interpreted as the \(l\)-step data-driven state transition matrix. Based on (\ref{equ:euq_g}), we can derive the general form of the data-driven MPC for trajectory planning:
\begin{equation}
    \begin{aligned}
    & \min_{(u,y)} \sum_{k=1}^{l} \left(\left\| u_k \right\|_R^2+ \left\| y_k - y_k^{\text{des}} \right\|_Q^2   \right) \\
    & \text{s.t.} \quad y = G \begin{bmatrix} u_{\text{ini}} \\ y_{\text{ini}} \\ u \end{bmatrix}, \\
    & \quad \quad \quad u_k \in \mathbf{U}, \; y_k \in \mathbf{Y}, \quad k = 1, \ldots, l.
    \end{aligned}
    \label{equ:euq_youhua}
\end{equation}

\subsection{Data-Driven Motion Planner}
\label{subsec:costfunction}
Our objective is to utilize data-driven MPC to leverage historical input-output data, enabling the FCRA to track a target trajectory in joint space. The trajectory is obtained by quintic polynomial interpolation in Sec.~\ref{subsec:Quintic polynomial interpolation}. At time \( t \), the available data are \( u_{t-N: t-1}^d \) and \( y_{t-N: t-1}^d \). Optimal control for position planning is achieved by solving the following \( l \)-step predictive control problem, which is formulated as a strictly convex quadratic programming problem:
\begin{subequations}\label{eq:main}
    \begin{align}
        \underset{(u, y)}{\text{min}}  & \|y_l- y^{ter}\|_S^2 + \sum_{k=1}^{l} (\|u_k - u^{tar}_k\|_R^2 + \|y_k - y^{tar}_k\|_Q^2 \nonumber \\ 
        & + \|u_{k+1}- u_{k}\|_F^2 + \|y_{k+1}- y_{k}\|_P^2) \label{eq:obj}\tag{\ref*{eq:main}a}\\   
        \text{s.t.} \quad & y = G \begin{bmatrix} u_{\text{ini}} \\ y_{\text{ini}} \\ u \end{bmatrix}, \label{eq:dynamics}\tag{\ref*{eq:main}b}\\
        & u_k \in \mathbf{U}, \; y_k \in \mathbf{Y}, \quad k = 1, \ldots, l, \label{eq:constraints}\tag{\ref*{eq:main}c}\\
        & \Delta y_{min}\leq y_{k+1}-y_k\leq \Delta y_{max}, \quad k = 1, \ldots, l. \label{eq:constraint7}\tag{\ref*{eq:main}d}
    \end{align}
\end{subequations}
        where $\left\| \mathbf{x} \right\| _{\mathbf{R}}\coloneqq \sqrt{\mathbf{x}^{\top}\mathbf{Rx}}$. ${R}$, ${F}\in \mathbb{R} ^{m\times m}$,
        ${Q}$, ${S}$, ${P} \in \mathbb{R} ^{c\times c}$ are positive definite.        $u_k$ and $y_k$ are the optimal variables in each step, corresponding to motor velocities and joint angles, respectively.
        ${u}^{tar}_{k}$ and ${y}^{tar}_{k}$ are the target values to be optimized for each step's input and output, serving as reference values.
        ${y}^{ter}$ is the terminal target value, representing the ultimate goal of the entire optimization process.
        The terms $u_{k+1} - u_{k}$ and $y_{k+1} - y_{k}$ represent the angular differences between consecutive steps for the motors and joints. 

        In the cost function, the term $\|u_k - u^{tar}_k\|_R^2$ promises the difference between motor velocity and target velocity in the $k$ step, while $\|y_k - y^{tar}_k\|_Q^2$ promises the difference between joint angle and target angle in the $k$ step. $\|y_l- y^{ter}\|_S^2$ promises the difference between the target angle in the $l$ step and the terminal target angle. $\left\| u_{k+1}-u_k \right\| _{{F}}^{2} $ and $\left\| y_{k+1}-y_k \right\| _{{P}}^{2}$ ensure the minimization of velocity and acceleration for motors and joints. 
        
        In the constraints, (\ref{eq:dynamics}) ensure that the predicted trajectory is consistent with the collected data and represents the model. (\ref{eq:constraints}) and (\ref{eq:constraint7}) respectively ensure that the predicted trajectory and the change in the output are within the upper and lower bounds.
        
        We formulate the motion planning problem and solve them using the OSQP\cite{stellato2020osqp} solver.

\subsection{Data Search Algorithm (DSA)}
\label{subsec:Data Search Algorithm}

In the actual system, we collect \( N_s \) data points that represent various operating conditions, such as the application of different loads to the system's end effector. However, using all these data points for computation would result in an extremely slow solution speed. Therefore, we divide the data into \( M \) groups, each containing \( N \) data points such that \( N_s = M \times N \). Additionally, we aim to fully leverage the system's characteristics to select appropriate data sequences and inverse MLP model references rapidly. This approach facilitates the accurate construction of matrices in subsequent steps, thereby reducing the solution time.

 We sample the current state of the system and select the data sequence to construct the \( G \) matrix described in Sec.~\ref{subsec:Construction of Data-Driven MPC}. We define \(\{u^d_{1:N,j}, y^d_{1:N,j}\}\) as the collected data sequences, where \( j \in \{1, \ldots, M\} \) denotes the index. The current sample data sequence is denoted as \(\{u^d_{L,th}, y^d_{L,th}\}\), which is sampled during the operation of the control and planning system. Here, \(N\) represents the length. We aim to select the best set from the \(M\) sets.

Following the paradigm of (\ref{equ:theorem1}), we compute the following expression:
\begin{equation}
    \begin{aligned}
    & \quad \quad \quad \quad \quad \quad \underset{j}{\text{min}} \| K_j \|^2 \\
    & \text{s.t.} 
    \begin{bmatrix}
    \mathcal{H}_L(u^d_ {1:N,j}) \\
    \mathcal{H}_L(y^d_ {1:N,j})
    \end{bmatrix}
    K_j =
    \begin{bmatrix}
    u_{L,th}^d \\
    y_{L,th}^d + \sigma_{L,th}
    \end{bmatrix},\\
    & \quad \quad \quad \quad \quad \quad  \sigma_{L,th} \subset \boldsymbol{\sigma}.
    \end{aligned}
    \label{equ:DSA}
\end{equation}
where $K_{j}$ is the column vector parameter to be determined, and $\sigma_{L,th}$ represents the slack variables introduced to address data noise issues. Using minimizing \(\|K_{j}\|^2\), the minimum value corresponds to the data sequence \(\{u^d_{1:N,be}, y^d_{1:N,be}\}\), which is identified as the most suitable. This selected data sequence $\{u^d_{1:N,be}, y^d_{1:N,be}\}$ is then used to construct the $G$ matrix. Furthermore, the minimum $\|K_{j}\|$ can be a criterion to select the corresponding inverse MLP reference model as input reference.

\subsection{Equivalent Model Reference}
\label{subsec:Equivalent Model Reference}
The collected data are used to fit the system model. Specifically, \( M \) system models with \(3n\) inputs and \(2n\) outputs are constructed using \(N_s\) input-output pairs of motor rotation angles and joint encoder values. 
The forward model is used to test the approach. The inverse model enables the deduction of motor rotation angles from joint angles, which is essential to incorporate motor speed as a tracking target in the velocity tracking cost. This process is illustrated in the MLP model fitting kinematics section depicted in Fig.~\ref{fig_framework}.
When the optimal \(\{u^d_{1:N,be}, y^d_{1:N,be}\}\) is obtained in Sec.~\ref{subsec:Data Search Algorithm}, the corresponding inverse MLP model will be selected from the \(M\) groups.

A three-layer perceptron is employed to fit both the forward and inverse models, using the root mean square error (RMSE) as the loss function and the Adam optimizer for training. The dataset is split into a training set, a validation set and a test set at a ratio of 8:1:1. The random seed is set at 42. The model hyperparameters are set as follows: a learning rate of 0.001, a batch size of 128, 100 training epochs, and a hidden layer width of 256 for each of the three layers, with real-time saving of the optimal model.

\subsection{Quintic Polynomial Interpolation}
\label{subsec:Quintic polynomial interpolation}
We use a quintic polynomial interpolation method for the FCRA to obtain a continuous reference trajectory only when the target of the system is a single point. The trajectory planning employs a fifth-order polynomial \( q(t) = a_0 + a_1 t + a_2 t^2 + a_3 t^3 + a_4 t^4 + a_5 t^5 \) to ensure smooth motion with continuous position, velocity and acceleration at both start (\( t=0 \)) and end (\( t=T \)) points. By setting boundary conditions of zero initial/final velocities (\( v_0 = v_T = 0 \)) and accelerations (\( \alpha_0 = \alpha_T = 0 \)), the six coefficients \( a_0, \ldots, a_5 \) are uniquely determined by solving the system.
The trajectory's continuity and differentiability make it ideal for high-precision tasks like drawing, where discretizing \( q(t) \) into time steps generates a seamless path, ensuring stable control and artifact-free curves.

\section{Experiments}

\begin{figure}[t]
    \centering
    \includegraphics[width=8.65cm]{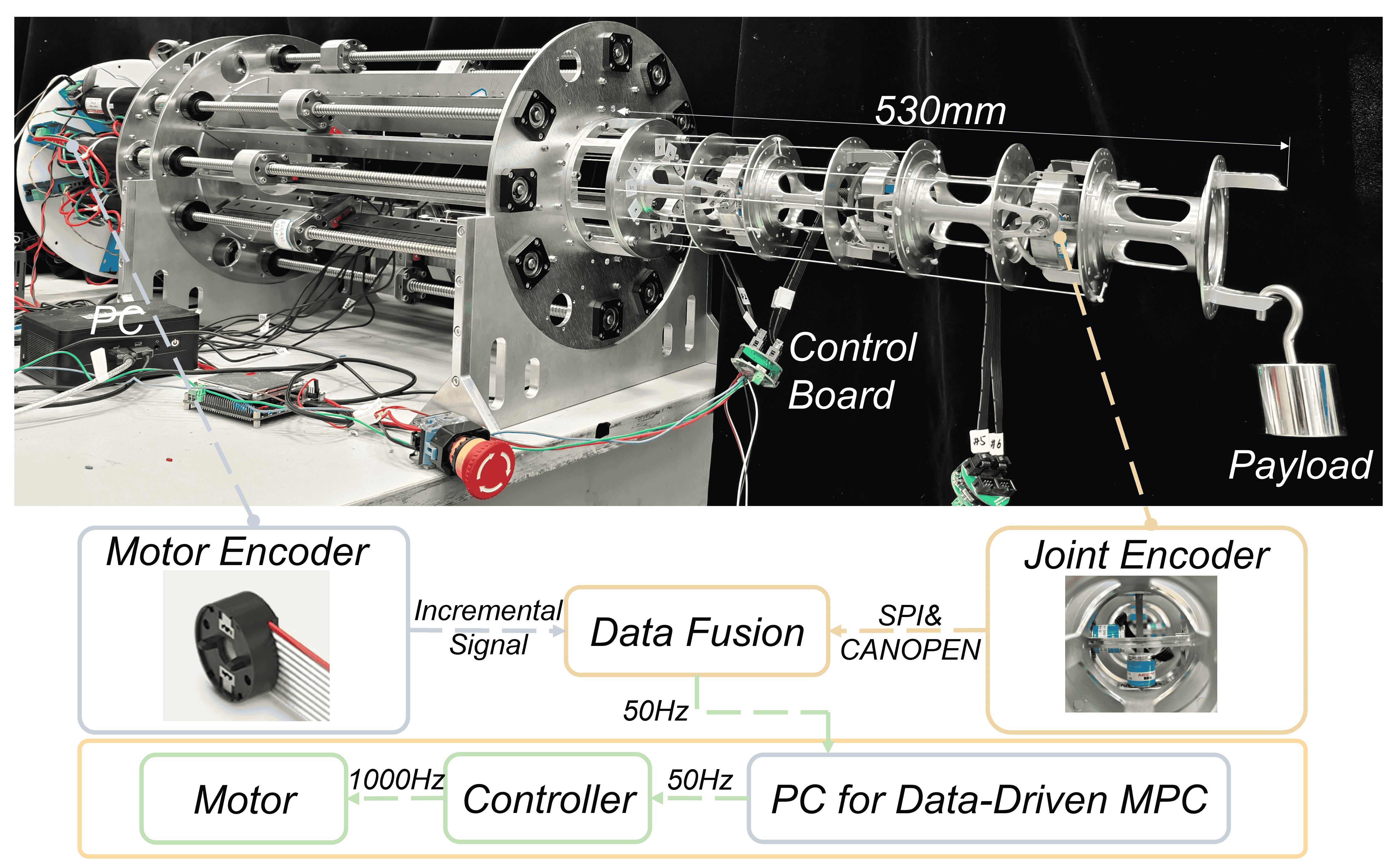}
    \caption{
        The real-world setup of the FCRA, with the number of segments equal to three, is equipped with motor encoders and joint encoders. The data are processed on the control board and transmitted to the PC for Data-Driven MPC computation, with the optimized results being sent back to the motor for control.
    }
    \label{fig:real_picture}
\end{figure}

For the FCRA discussed in Fig.~\ref{fig:biaoshi}, we construct a real-world experimental platform, as shown in Fig.~\ref{fig:real_picture}, for the case where \( n=3\),  \( m=9\), \( c=6 \). The sensor system comprises motor encoders and joint encoders. Data are processed on the control board and transmitted to the PC for data-driven MPC computation, with the optimized results being sent back to the motor for control. The entire process is coordinated via ROS, with the information flow operating at 50 Hz, encompassing both MPC computation and feedback reception. Based on this platform, we perform a series of experiments to validate the effectiveness of the method depicted in Fig.~\ref{fig_framework}. 

\subsection{DSA Verification Experiment}
\label{subsec:DSA Experimental Verification}
The \( N_s \) collected data points, each sampled at intervals of 0.02 seconds, corresponding to \( M = 6 \) and \( N = 120000 \). The value of \( N \) is restricted to (\ref{equ:simu_hankel}).
Now, we verify the feasibility of the DSA described in Sec.~\ref{subsec:Data Search Algorithm}. During the data collection phase, loads ranging from 0 to 2.5 kg, with increments of 0.5 kg, are applied to the FCRA end-effector, as shown in Fig.~\ref{fig:real_picture}. Subsequently, the current system collects three separate payload datasets, which are then substituted into (\ref{equ:DSA}). The results are presented in Table \ref{table:DSA_result}. The optimal data pair \( \{u^d_{1:N,be}, y^d_{1:N,be}\} \) corresponds to the index \( j \) that minimizes \( \|K_{j}\| \).

Clearly, the minimum value of \( \|K_j\| \) corresponds to the load closest to the actual load.
Moreover, directly using the average time to process all data points \( N_s \) is approximately 19 ms, while solving using the selected data groups takes approximately 4 ms each step, resulting in a reduction of nearly 80\%.
The results demonstrate that the DSA can effectively identify the data pairs that best match the current state of the system, thereby reducing the solution time for data-driven MPC planning.

\begin{table}[!h]
    \centering
    \caption{The corresponding $\|K_j\|$ values for each load}
    \label{table:DSA_result}
    \resizebox{\linewidth}{!}{ 
    \begin{tabular}{|c|c|c|c|c|c|c|c|}
        \hline
        $Load$ & 
        $Weight(kg)$ &
        $0$ & 
        $0.5$ &
        $1$ &
        $1.5$ & 
        $2$ & 
        $2.5$ \\ 
        \hline 
        $Payload 1$ &\(0.032\) & \(\textbf{683.55}\) & \(689.08\) & \(701.02\) & \(740.86\) & \(747.86\) & \(795.13\)\\ 
        $Payload 2$ &\(1.092\) & \(699.77\) & \(695.31\) & \(\textbf{685.05}\) & \(741.11\) & \(748.13\) & \(790.24\)\\ 
        $Payload 3$ &\(2.088\) & \(780.81\) & \(746.22\) & \(738.33\) & \(702.11\) & \(\textbf{680.16}\) & \(686.02\)\\ 
        \hline 
    \end{tabular}
    }
\end{table}

\subsection{Simulation-based Hyperparameter Tuning Analysis}
\label{subsec:Simulation-based Hyperparameter Tuning Analysis}
\begin{figure}[t]
    \centering
    \includegraphics[width=8.65cm]{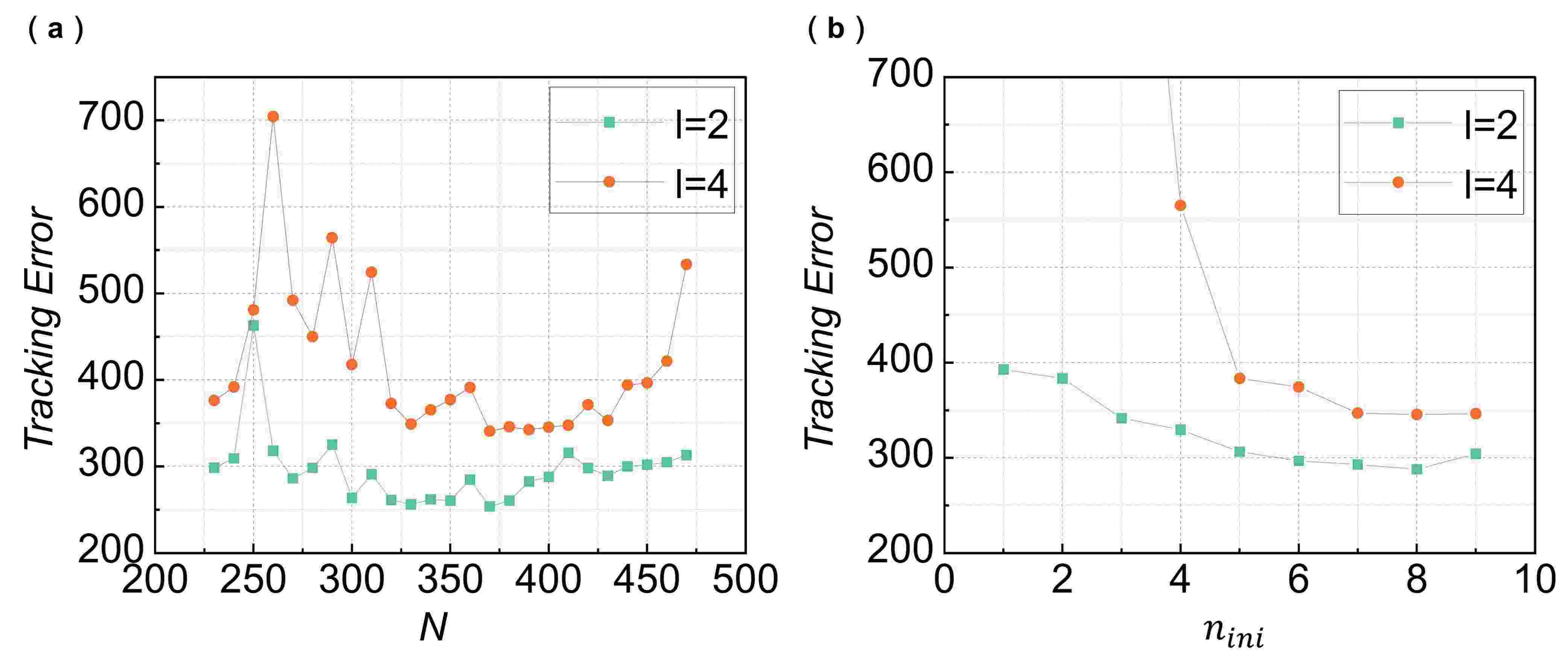}
    \caption{
        The relationship between hyperparameters and tracking error, where \( l \) denotes the prediction horizon. (a) illustrates the impact of $N$ on the overall tracking error, where $n_{\text{ini}}=8$, while (b) depicts the effect of $n_{\text{ini}}$ on the tracking error, where $N=401$.
    }
    \label{fig:simulation_hyperparameter}
\end{figure}

 The constructed MLP models are first used to perform hyperparameter tuning analysis. For nonlinear systems with significant noise, the analysis focuses on selecting the most appropriate hyperparameters to construct the matrix in (\ref{equ:hankel}), thus minimizing the tracking error and ensuring that the specific minimum values satisfy the conditions. Since the construction of the matrix is mainly influenced by \( N \) and \( n_{\text{ini}} \), the hyperparameters include the estimation horizon \( n_{\text{ini}} \) and the total length of the collected data \( N \).
 Fig.~\ref{fig:simulation_hyperparameter} shows how they affect the tracking error.

The data-driven MPC planning in Fig.~\ref{fig_framework} is activated. 
The robotic arm is commanded to move from $E_s$ = (340mm, 150mm, -40mm) to $E_e$ = (360mm, 20mm, -90mm). 
The closed-loop tracking error is measured as 
\begin{equation}
    \frac{1}{c}\sum_{t=0}^{T_a-1} \left\| y_t - y^{\text{ter}} \right\|_2^2
    \label{equ:error}
\end{equation}
The end time of the step trajectory is $T_a = 1200$, equivalent to $24$ seconds in real-world scenarios.

The key conclusion from Fig.~\ref{fig:simulation_hyperparameter}(a) is that the tracking error significantly improves when the Hankel matrix becomes at least square, particularly when \( l=4 \). However, this property no longer holds when the number of columns is less than the number of rows. This conclusion can be summarized by the following inequality, which indicates that \( N \) is needed to ensure the Hankel matrix is square:
\begin{equation}
    N \geq \max\{(m+1)(n_{\text{ini}} + l + h) , (m + c + 1)(n_{\text{ini}} + l) \}
    \label{equ:simu_hankel}
\end{equation}
where \( h=15 \) denotes the number of states.

In Fig.~\ref{fig:simulation_hyperparameter}(b), increasing \( n_{\text{ini}} \) leads to a notable reduction in tracking error. When \( l=4\), the tracking performance improves further. Under noisy conditions, increasing \( n_{\text{ini}} \) provides better initial condition estimation.

\subsection{Position Repeatability}
\label{subsec:Position Repeatability}

\begin{figure}[t]
    \centering
    \includegraphics[width=3.65cm]{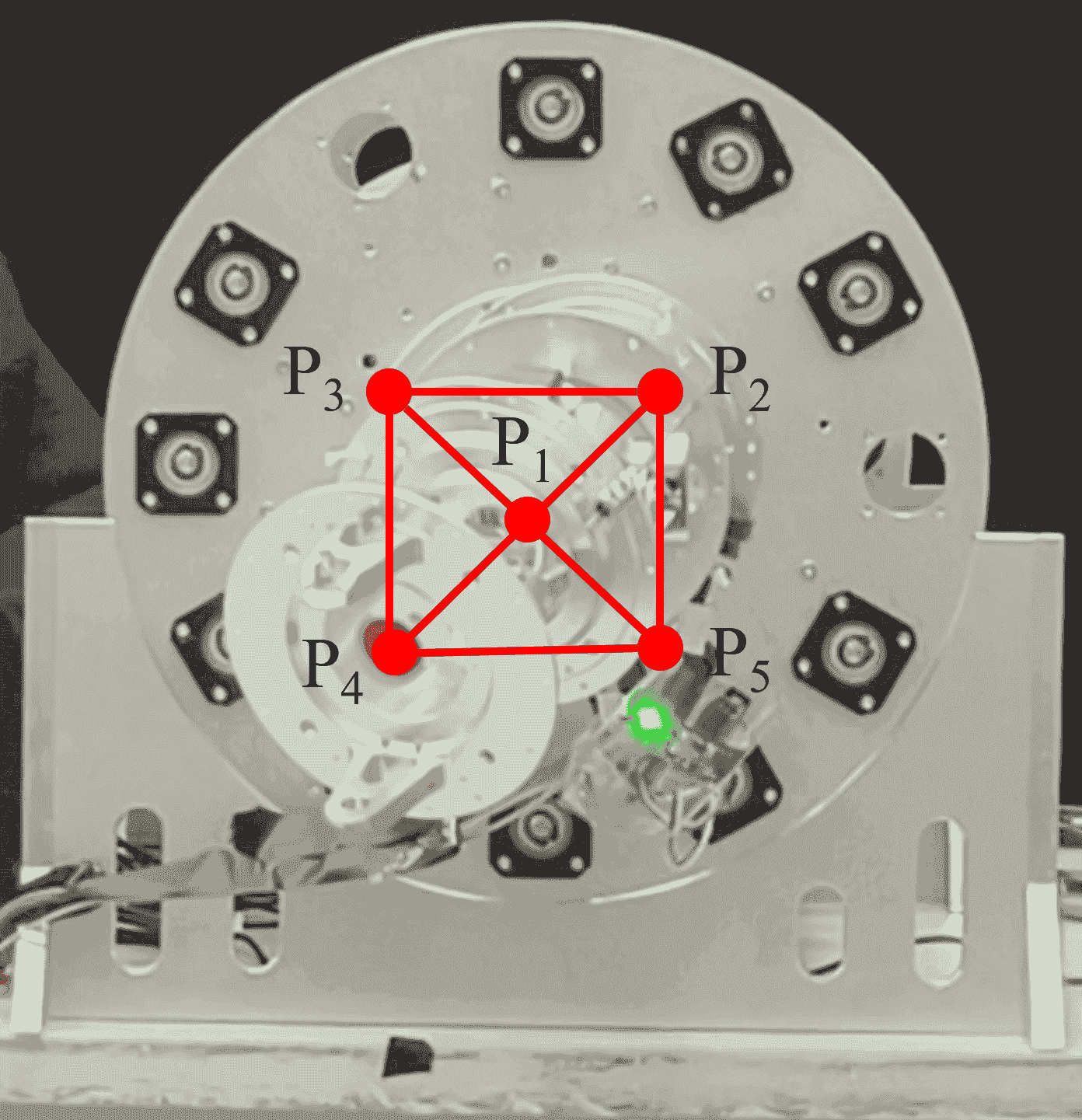}
    \caption{
        The repeatability test has five pre-defined target positions ($P_1, P_2, P_3, P_4, P_5$). The FCRA moves between these positions and repeats thirty times.
    }
    \label{fig:wudiandingwei}
\end{figure}

To evaluate the repeatability of the data-driven MPC for the FCRA, we conduct calibration experiments on the apparatus, as shown in Fig.~\ref{fig:wudiandingwei}. Five predefined target positions are established, and the FCRA moves between $P_1$, $P_2$, $P_3$, $P_4$, and $P_5$, repeating the process thirty times. The computation is carried out through the forward kinematics from the joints to the operational space. After performing inverse kinematics at these points to track the joint angles, the average distance (mean), standard deviation (STD.DEV), and 3-sigma distance between the mean and recorded positions are summarized in Table.~\ref{table:Position Repeatability Test Results}. The average position error of the FCRA is 2.070 mm, which is less than the 4.9 mm of the Twist Snake \cite{10160995}, thereby demonstrating the effectiveness and accuracy of the repeatability of the algorithm. 

\begin{table}[!htbp]
\centering
\caption{Results of Position Repeatability Test}
    \label{table:Position Repeatability Test Results}
        \resizebox{\linewidth}{!}{ 
\begin{tabular}{|c|c|c|c|}
\hline
\textbf{Pose} & {Mean(mm)} & {STD.DEV.(mm)} & {3-SIGMA(mm)} \\ \hline
$P_1$ & 2.820 & 2.132 & 9.216 \\ 
$P_2$ & 1.428 & 1.496 & 5.916 \\ 
$P_3$ & 1.824 & 1.244 & 5.556 \\ 
$P_4$ & 2.460 & 2.640 & 10.380 \\ 
$P_5$ & 1.816 & 1.956 & 7.684 \\ 
\textbf{Average} & \textbf{2.070} & \textbf{1.894} & \textbf{7.750} \\ \hline
\end{tabular}
    }
\end{table}

The 3-sigma distance is 7.750 mm. We attribute the error primarily to the slight elongation of the cables, especially when the number of test cycles is large and the operating time is extended. This elongation can cause minor slackening in the cables, thereby amplifying the system error. Given the cable length of over one meter, this effect is notable.

\subsection{Tracking Performance}
\label{subsec:Tracking performance on the experimental platform}
\begin{figure}[t]
    \centering
    \includegraphics[width=8.65cm]{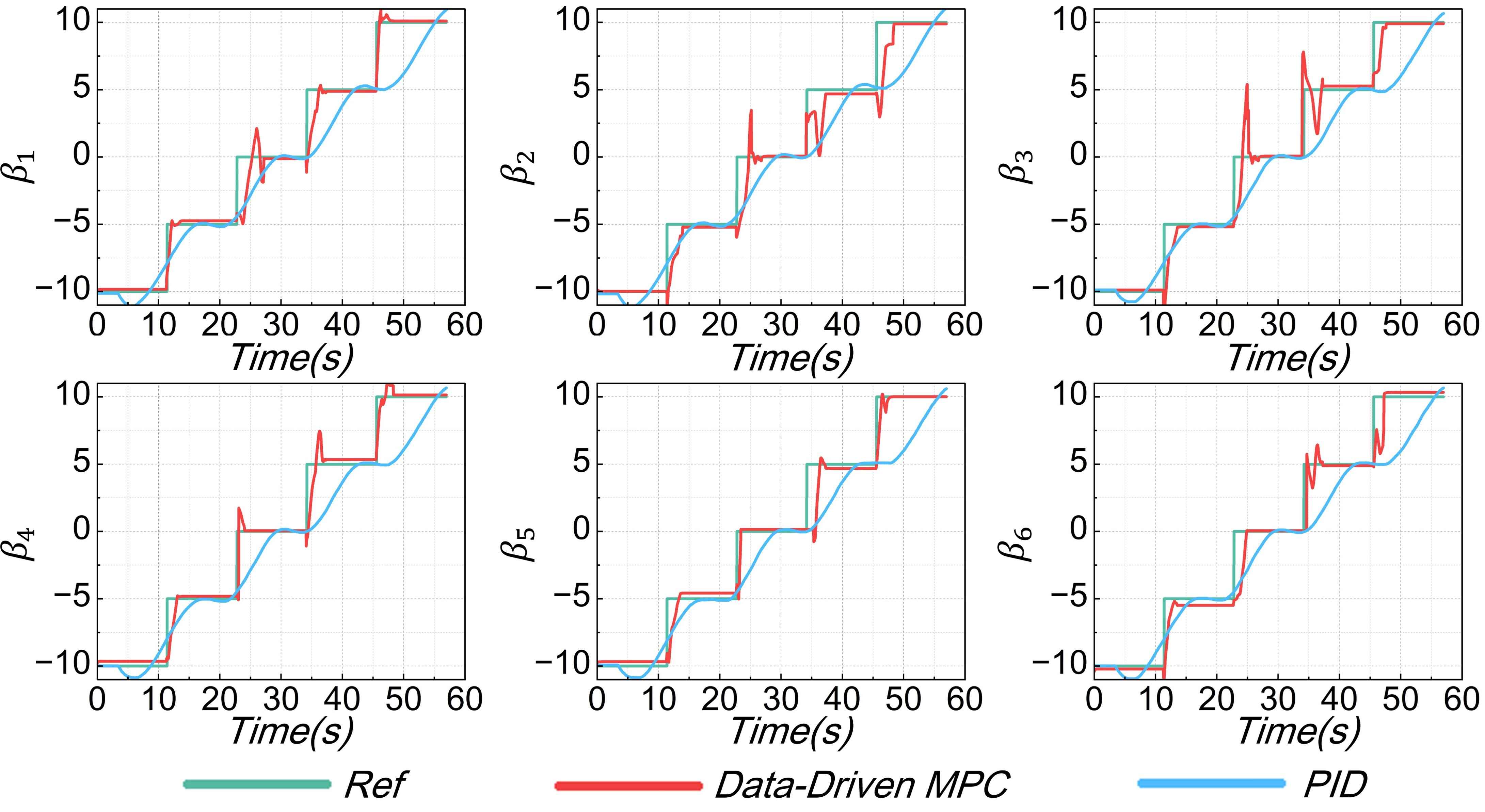}
    \caption{
This illustrates the single-point tracking performance on the FCRA, where the red line denotes the actual value of the data-driven MPC, the blue line denotes the actual value of the PID method, and the green line denotes the reference value.
    }
    \label{fig:jieyue}
\end{figure}

To verify the single-point tracking performance of the data-driven MPC and compare it with the PID method \cite{li2022design}, we conduct comparative experiments on the FCRA. As shown in Fig.~\ref{fig:jieyue} and Table.~\ref{table:Average Tracking Error at Each Target Position}, the average error of the PID method is 1.418°, whereas the average error of the data-driven MPC is 0.541°. The parameters in the data-driven MPC planner are taken to be \(l =6 \), \(n_{\text{ini}}=2\), $Q = 10000I_{1X6}$, $R = 70I_{1X9}$, $S = 0.01I_{1X6}$, $F = 0.01I_{1X9}$, and $P = 0.1I_{1X6}$. The results indicate that planning consistently achieves satisfactory tracking performance, even when the target position undergoes sudden changes. It can rapidly follow the target position point. In contrast, the PID method exhibits significant lag, possibly due to the hysteresis characteristics of the cable, which requires more time to catch up with the target point. 
Therefore, the data-driven MPC outperforms the PID method in terms of single-point tracking error. This highlights the limitations of the PID method when adapting to complex cable-driven systems, whereas the data-driven MPC is supported by a more robust theoretical foundation.

\begin{figure}[t]
    \centering
    \includegraphics[width=8.65cm]{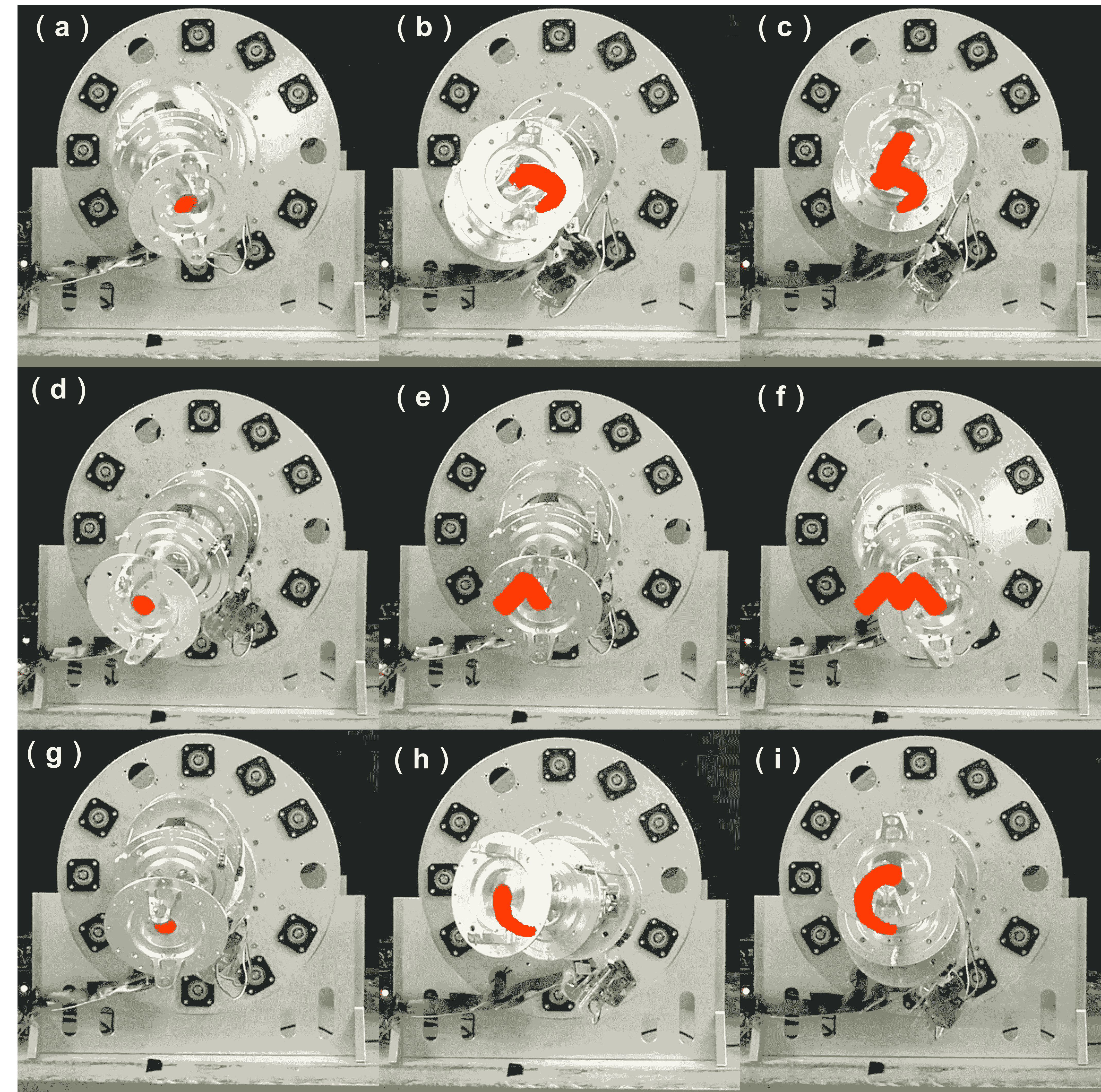}
    \caption{
        Intermediate steps of ``SMC". (a)-(c) present the trajectory tracking results for the letter `S'. (d)-(f) show the trajectory tracking results for the letter `M'. (g)-(i) illustrate the trajectory tracking results for the letter `C'.
    }
    \label{fig:SMC_fenduan}
\end{figure}

Furthermore, the experiment is extended to track the trajectory of the letters ``SMC" as depicted in Fig.~\ref{fig:SMC}. Without relying on modeling, we successfully track the trajectory of the three letters in a plane with clear and accurate results. The intermediate steps to draw these letters are shown in Fig.~\ref{fig:SMC_fenduan}.

These results clearly demonstrate the capability of the proposed model-free data-driven MPC method to track complex trajectories on the experimental platform accurately. This highlights the robustness and adaptability of planning in addressing real-world scenarios with diverse trajectory requirements.

\begin{table}[!htbp]
\centering
\caption{Average Tracking Error at Each Target Position}
    \label{table:Average Tracking Error at Each Target Position}
        \resizebox{\linewidth}{!}{ 
\begin{tabular}{|c|c|c|}
\hline
\textbf{Target Position} & {Error of PID(°)} & {Error of Data-driven MPC(°)}  \\ \hline
$TT_1$ & 0.561 & 0.205  \\ 
$TT_2$ & 0.616 & 0.612  \\ 
$TT_3$ & 1.208 & 0.670  \\ 
$TT_4$ & 1.893 & 0.715  \\ 
$TT_5$ & 2.814 & 0.502  \\ 
\textbf{Average} & \textbf{1.418} & \textbf{0.541}  \\ \hline
\end{tabular}
    }
\end{table}

\section{Conclusion and Future Work}
\label{sec:conclusion}

This study successfully overcame the modeling challenges associated with the FCRA by developing an enhanced MPC method that relies exclusively on input-output data to predict system behavior. The DSA significantly reduced the computation time for each optimization iteration from about 19 to 4 ms. Additionally, the study explored the impact of hyperparameters on error accuracy and experimentally verified the proposed algorithm on an FCRA testbed. The findings indicated that the average positioning accuracy across the five test points was approximately 2.070 mm. Furthermore, compared to the PID control method, which exhibited an average tracking error of 1.418°, the data-driven MPC achieved a substantially lower average tracking error of 0.541°.
Future work will focus on more complex system control tasks, such as robotic arm grasping\cite{10802005,10802595}, and further optimization of the proposed algorithm to enhance its adaptability and robustness in dynamic environments.


\bibliographystyle{IEEEtran}
\bibliography{IEEEabrv, bib/bibliography}

\begin{thebibliography}{10}
\providecommand{\url}[1]{#1}
\csname url@samestyle\endcsname
\providecommand{\newblock}{\relax}
\providecommand{\bibinfo}[2]{#2}
\providecommand{\BIBentrySTDinterwordspacing}{\spaceskip=0pt\relax}
\providecommand{\BIBentryALTinterwordstretchfactor}{4}
\providecommand{\BIBentryALTinterwordspacing}{\spaceskip=\fontdimen2\font plus
\BIBentryALTinterwordstretchfactor\fontdimen3\font minus \fontdimen4\font\relax}
\providecommand{\BIBforeignlanguage}[2]{{%
\expandafter\ifx\csname l@#1\endcsname\relax
\typeout{** WARNING: IEEEtran.bst: No hyphenation pattern has been}%
\typeout{** loaded for the language `#1'. Using the pattern for}%
\typeout{** the default language instead.}%
\else
\language=\csname l@#1\endcsname
\fi
#2}}
\providecommand{\BIBdecl}{\relax}
\BIBdecl

\bibitem{gravagne2002manipulability}
I.~A. Gravagne and I.~D. Walker, ``Manipulability, force, and compliance analysis for planar continuum manipulators,'' \emph{IEEE Transactions on Robotics and Automation}, vol.~18, no.~3, pp. 263--273, 2002.

\bibitem{xu2008investigation}
K.~Xu and N.~Simaan, ``An investigation of the intrinsic force sensing capabilities of continuum robots,'' \emph{IEEE Transactions on Robotics}, vol.~24, no.~3, pp. 576--587, 2008.

\bibitem{porto2019position}
R.~A. Porto, F.~Nageotte, P.~Zanne, and M.~de~Mathelin, ``Position control of medical cable-driven flexible instruments by combining machine learning and kinematic analysis,'' in \emph{2019 international conference on robotics and automation (ICRA)}.\hskip 1em plus 0.5em minus 0.4em\relax IEEE, 2019, pp. 7913--7919.

\bibitem{baek2020hysteresis}
D.~Baek, J.-H. Seo, J.~Kim, and D.-S. Kwon, ``Hysteresis compensator with learning-based hybrid joint angle estimation for flexible surgery robots,'' \emph{IEEE Robotics and Automation Letters}, vol.~5, no.~4, pp. 6837--6844, 2020.

\bibitem{luo2025d3}
H.~Luo, J.~Xu, S.~Li, H.~Liang, Y.~Chen, C.~Xia, and X.~Wang, ``D3-arm: High-dynamic, dexterous and fully decoupled cable-driven robotic arm,'' in \emph{2025 IEEE International Conference on Robotics and Automation (ICRA)}.\hskip 1em plus 0.5em minus 0.4em\relax IEEE, 2025.

\bibitem{li2022design}
W.~Li, W.~Xu, B.~Lin, and L.~Yan, ``Design, kinematics and control of a modular cable-driven manipulator for fine manipulation,'' in \emph{2022 IEEE International Conference on Robotics and Biomimetics (ROBIO)}.\hskip 1em plus 0.5em minus 0.4em\relax IEEE, 2022, pp. 833--838.

\bibitem{berberich2020data}
J.~Berberich, J.~K{\"o}hler, M.~A. M{\"u}ller, and F.~Allg{\"o}wer, ``Data-driven model predictive control with stability and robustness guarantees,'' \emph{IEEE Transactions on Automatic Control}, vol.~66, no.~4, pp. 1702--1717, 2020.

\bibitem{coulson2021distributionally}
J.~Coulson, J.~Lygeros, and F.~D{\"o}rfler, ``Distributionally robust chance constrained data-enabled predictive control,'' \emph{IEEE Transactions on Automatic Control}, vol.~67, no.~7, pp. 3289--3304, 2021.

\bibitem{bongard2022robust}
J.~Bongard, J.~Berberich, J.~K{\"o}hler, and F.~Allg{\"o}wer, ``Robust stability analysis of a simple data-driven model predictive control approach,'' \emph{IEEE Transactions on Automatic Control}, vol.~68, no.~5, pp. 2625--2637, 2022.

\bibitem{wang2018practical}
Y.~Wang, F.~Yan, K.~Zhu, B.~Chen, and H.~Wu, ``Practical adaptive integral terminal sliding mode control for cable-driven manipulators,'' \emph{IEEE Access}, vol.~6, pp. 78\,575--78\,586, 2018.

\bibitem{hannan2003kinematics}
M.~W. Hannan and I.~D. Walker, ``Kinematics and the implementation of an elephant's trunk manipulator and other continuum style robots,'' \emph{Journal of robotic systems}, vol.~20, no.~2, pp. 45--63, 2003.

\bibitem{camarillo2009task}
D.~B. Camarillo, C.~R. Carlson, and J.~K. Salisbury, ``Task-space control of continuum manipulators with coupled tendon drive,'' in \emph{Experimental Robotics: The Eleventh International Symposium}.\hskip 1em plus 0.5em minus 0.4em\relax Springer, 2009, pp. 271--280.

\bibitem{qin1997overview}
S.~J. Qin and T.~A. Badgwell, ``An overview of industrial model predictive control technology,'' in \emph{AIche symposium series}, vol.~93, no. 316.\hskip 1em plus 0.5em minus 0.4em\relax New York, NY: American Institute of Chemical Engineers, 1971-c2002., 1997, pp. 232--256.

\bibitem{jian2023dynamic}
Z.~Jian, Z.~Yan, X.~Lei, Z.~Lu, B.~Lan, X.~Wang, and B.~Liang, ``Dynamic control barrier function-based model predictive control to safety-critical obstacle-avoidance of mobile robot,'' in \emph{2023 IEEE International Conference on Robotics and Automation (ICRA)}.\hskip 1em plus 0.5em minus 0.4em\relax IEEE, 2023, pp. 3679--3685.

\bibitem{chen2023quadruped}
Y.~Chen, Z.~Xu, Z.~Jian, G.~Tang, L.~Yang, A.~Xiao, X.~Wang, and B.~Liang, ``Quadruped guidance robot for the visually impaired: A comfort-based approach,'' in \emph{2023 IEEE International Conference on Robotics and Automation (ICRA)}.\hskip 1em plus 0.5em minus 0.4em\relax IEEE, 2023, pp. 12\,078--12\,084.

\bibitem{fawcett2022toward}
R.~T. Fawcett, K.~Afsari, A.~D. Ames, and K.~A. Hamed, ``Toward a data-driven template model for quadrupedal locomotion,'' \emph{IEEE Robotics and Automation Letters}, vol.~7, no.~3, pp. 7636--7643, 2022.

\bibitem{coulson2019data}
J.~Coulson, J.~Lygeros, and F.~D{\"o}rfler, ``Data-enabled predictive control: In the shallows of the deepc,'' in \emph{2019 18th European Control Conference (ECC)}.\hskip 1em plus 0.5em minus 0.4em\relax IEEE, 2019, pp. 307--312.

\bibitem{berberich2021data}
J.~Berberich, J.~K{\"o}hler, M.~A. M{\"u}ller, and F.~Allg{\"o}wer, ``Data-driven model predictive control: closed-loop guarantees and experimental results,'' \emph{at-Automatisierungstechnik}, vol.~69, no.~7, pp. 608--618, 2021.

\bibitem{schmitt2023data}
L.~Schmitt, J.~Beerwerth, M.~Bahr, and D.~Abel, ``Data-driven predictive control with online adaption: Application to a fuel cell system,'' \emph{IEEE Transactions on Control Systems Technology}, 2023.

\bibitem{huang2019data}
L.~Huang, J.~Coulson, J.~Lygeros, and F.~D{\"o}rfler, ``Data-enabled predictive control for grid-connected power converters,'' in \emph{2019 IEEE 58th Conference on Decision and Control (CDC)}.\hskip 1em plus 0.5em minus 0.4em\relax IEEE, 2019, pp. 8130--8135.

\bibitem{xiong2020comparison}
H.~Xiong, T.~Ma, L.~Zhang, and X.~Diao, ``Comparison of end-to-end and hybrid deep reinforcement learning strategies for controlling cable-driven parallel robots,'' \emph{Neurocomputing}, vol. 377, pp. 73--84, 2020.

\bibitem{yang2023manipulability}
H.~Yang, X.~Li, D.~Meng, X.~Wang, and B.~Liang, ``Manipulability optimization of redundant manipulators using reinforcement learning,'' \emph{Industrial Robot: the international journal of robotics research and application}, vol.~50, no.~5, pp. 830--840, 2023.

\bibitem{10160491}
H.~Zhang, H.~Liang, L.~Cong, J.~Lyu, L.~Zeng, P.~Feng, and J.~Zhang, ``Reinforcement learning based pushing and grasping objects from ungraspable poses,'' in \emph{2023 IEEE International Conference on Robotics and Automation (ICRA)}, 2023, pp. 3860--3866.

\bibitem{willems2005note}
J.~C. Willems, P.~Rapisarda, I.~Markovsky, and B.~L. De~Moor, ``A note on persistency of excitation,'' \emph{Systems \& Control Letters}, vol.~54, no.~4, pp. 325--329, 2005.

\bibitem{stellato2020osqp}
B.~Stellato, G.~Banjac, P.~Goulart, A.~Bemporad, and S.~Boyd, ``Osqp: An operator splitting solver for quadratic programs,'' \emph{Mathematical Programming Computation}, vol.~12, no.~4, pp. 637--672, 2020.

\bibitem{10160995}
K.~Tanaka and M.~Hamaya, ``Twist snake: Plastic table-top cable-driven robotic arm with all motors located at the base link,'' in \emph{2023 IEEE International Conference on Robotics and Automation (ICRA)}, 2023, pp. 7345--7351.

\bibitem{10802005}
Y.~H. Xie, W.~J. Lv, X.~Y. Zhang, Y.~H. Chen, and L.~Zeng, ``Parametricnet++: A 6dof pose estimation network with sparse keypoint recovery for parametric shapes in stacked scenarios,'' in \emph{2024 IEEE/RSJ International Conference on Intelligent Robots and Systems (IROS)}, 2024, pp. 7181--7188.

\bibitem{10802595}
D.-T. Huang, E.-T. Lin, L.~Chen, L.-F. Liu, and L.~Zeng, ``Sd-net: Symmetric-aware keypoint prediction and domain adaptation for 6d pose estimation in bin-picking scenarios,'' in \emph{2024 IEEE/RSJ International Conference on Intelligent Robots and Systems (IROS)}, 2024, pp. 2747--2754.

\end{thebibliography}
\end{document}